\documentclass[10pt,twocolumn,letterpaper]{article}

\usepackage{geometry} 
\usepackage{iccv}
\usepackage{times}
\usepackage{epsfig}
\usepackage{graphicx}
\usepackage{amsmath}
\usepackage{amssymb}
\usepackage{url}
\usepackage{wrapfig}
\usepackage{footnote}
\usepackage{enumitem}
\usepackage{amsthm}
\usepackage{color,xcolor,ucs}
\newcommand*{\Scale}[2][4]{\scalebox{#1}{$#2$}}

\usepackage{algorithm}
\usepackage{algorithmic}
\usepackage{amsfonts}
\usepackage{wasysym}
\usepackage[ruled,vlined,algo2e]{algorithm2e}
\providecommand{\SetAlgoLined}{\SetLine}
\usepackage{tablefootnote}
\usepackage{multirow}

\def \c {\mathbf{c}}
\def \x {\mathbf{x}}
\def \w {\mathbf{w}}

\def \S {\mathcal{S}}

\DeclareMathOperator*{\argmin}{arg\,min}
\DeclareMathOperator*{\argmax}{arg\,max}

\newtheorem*{definition*}{Definition}
\newtheorem*{cor*}{Corollary}

\newcommand{\rmnum}[1]{\romannumeral #1}

\usepackage[pagebackref=true,breaklinks=true,letterpaper=true,colorlinks,bookmarks=false]{hyperref}

\iccvfinalcopy 

\def\iccvPaperID{1728} 

\ificcvfinal\fi
\begin{document}

\title{Zero-Shot Learning via Semantic Similarity Embedding}

\author{Ziming Zhang and Venkatesh Saligrama\\
Department of Electrical \& Computer Engineering, Boston University\\
{\tt\small \{zzhang14, srv\}@bu.edu}
}

\maketitle

\begin{abstract}

In this paper we consider a version of the zero-shot learning problem where seen class source and target domain data are provided. The goal during test-time is to accurately predict the class label of an unseen target domain instance based on revealed source domain side information (\eg attributes) for unseen classes. Our method is based on viewing each source or target data as a mixture of seen class proportions and we postulate that the mixture patterns have to be similar if the two instances belong to the same unseen class. This perspective leads us to learning source/target embedding functions that map an arbitrary source/target domain data into a same semantic space where similarity can be readily measured. We develop a max-margin framework to learn these similarity functions and jointly optimize parameters by means of cross validation. 
Our test results are compelling, leading to significant improvement in terms of accuracy on most benchmark datasets for zero-shot recognition.

\end{abstract}

\section{Introduction} 
While there has been significant progress in large-scale classification in recent years \cite{ILSVRCarxiv14}, lack of sufficient training data for every class and the increasing difficulty in finding annotations for a large fraction of data might impact further improvements. 

Zero-shot learning is being increasingly recognized as a way to deal with these difficulties. One version of zero shot learning is based on so-called source and target domains. Source domain is described by a {\it single} vector corresponding to each class based on side information such as {\em attributes} \cite{farhadi2009attribute,10.1109/TPAMI.2013.140,mensink2012metric,Parikh:2011:IBD:2191740.2191861,rohrbach2011largeScale}, {\em language words/phrases} \cite{Berg:2010:AAD:1886063.1886114,frome2013devise,socher2013zero}, or even {\em learned classifiers} \cite{yu2013designing}, which we assume can be collected easily. The target domain is described by a joint distribution of images/videos and labels \cite{10.1109/TPAMI.2013.140, wu2014zero}. During training time, we are given source domain attributes and target domain data corresponding to only a subset of classes, which we call seen classes. During test time, source domain attributes for unseen (\ie no training data provided) classes are revealed. The goal during test time is to predict for each target domain instance which of the seen/unseen classes it is associated with. 

\begin{figure}[t]
\centerline{\includegraphics[width=.9\columnwidth]{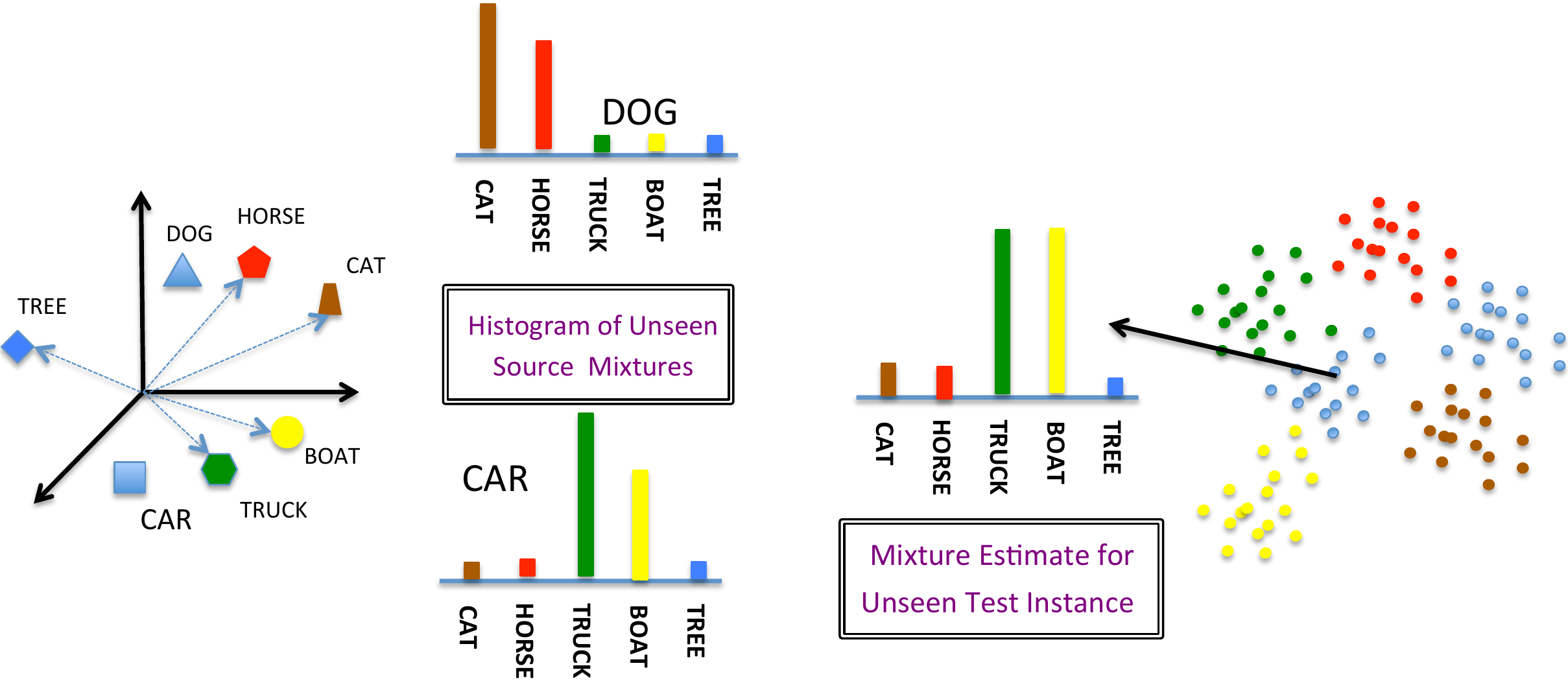}}
\caption{\footnotesize{Proposed method with source/target domain data displayed on the leftmost/rightmost figures respectively. Light blue corresponds to unseen classes and other colors depict seen class data. Light-blue data is unavailable during training. During test-time unseen source domain data is revealed along with an arbitrary unseen instance from target domain (light-blue) is presented and we are to identify its unseen class label. Each unseen class source domain data is expressed as a histograms of seen class proportions. Seen class proportions are estimated for the target instance and compared with each of the source domain histograms.}}
%
%
\label{fig:overview}
\vspace{-2mm}
\end{figure}

\noindent
{\bf Key Idea:} Our proposed method is depicted in Fig. \ref{fig:overview}. We view target data instances as arising from seen instances and attempt to express source/target data as a mixture of seen class proportions. Our algorithm is based on the postulate that if the mixture proportion from target domain is similar to that from source domain, they must arise from the same class. This leads us to learning source and target domain embedding functions using seen class data that map arbitrary source and target domain data into mixture proportions of seen classes.

We propose parameterized-optimization problems for learning {\em semantic similarity embedding} (SSE) functions from training data and jointly optimize predefined parameters using cross validation on held-out seen class data. Our method necessitates fundamentally new design choices requiring us to learn class-dependent feature transforms because components of our embedding must account for contribution of each seen class.
%
Our source domain embedding is based on subspace clustering literature \cite{vidal2010tutorial} that are known to be resilient to noise. 
%
%
%
Our target domain embedding is based on a margin-based framework using the intersection function or the rectified linear unit (ReLU) \cite{nair2010rectified}, which attempts to align seen class source domain data with their corresponding seen class target domain data instances. Finally, we employ a cross validation technique based on holding out seen class data and matching held-out seen classes to optimize parameters used in the optimization problems for source and target domain. In this way we jointly optimize parameters to best align mixture proportions for held-out seen classes and provide a basis for generalizing to unseen classes.
%
Results on several benchmark datasets for zero-shot learning demonstrate that our method significantly improves the current state-of-the-art results.

\noindent
{\bf Related Work:} Most existing zero-shot learning methods rely on predicting side information for further classification. \cite{palatucci2009zero} proposed a semantic (\ie attribute) output code
classifier which utilizes a knowledge base of semantic properties. \cite{10.1109/TPAMI.2013.140,wang2013unified} proposed several probabilistic attribute prediction methods. 
\cite{yu2013designing} proposed designing discriminative category-level attributes. \cite{mahajan2011joint} proposed an optimization formulation to learn source domain attribute classifiers and attribute vectors jointly. \cite{mensink2014costa} proposed learning the classifiers for unseen classes by linearly combining the classifiers for seen classes. \cite{akata2013label} proposed a label embedding method to embed each class into an attribute vector space. \cite{Akata2015,frome2013devise,norouziMBSSFCD14,socher2013zero} directly learned the mapping functions between the feature vectors in source and target domains with deep learning. Such methods may suffer from noisy (\eg missing or incorrectly annotated) side information or data bias, leading to unreliable prediction. 

Some recent work has been proposed to overcome some issues above. \cite{conf/nips/RohrbachES13} proposed a propagated semantic transfer method by exploiting unlabeled instances. \cite{embedding2014ECCV} discussed the projection domain shift problem and proposed a transductive multi-view embedding method. \cite{jayaraman2014unreliable} investigated the attribute unreliability issue and proposed a random forest approach. \cite{Romera-Paredes2015} proposed a simple method by introducing a better regularizer.


An important conceptual difference that distinguishes our method from other existing works such as \cite{akata2013label,Akata2015}, is that these methods can be interpreted as learning relationships between source attributes and target feature components (in the encoded space), while our method is based on leveraging similar class relationships (semantic affinities) in source and target domains, requiring class dependent feature transform. This leads to complex scoring functions, which cannot be simplified to linear or bilinear forms as in \cite{akata2013label,Akata2015}.

{\em Semantic similarity embedding (SSE)} is widely used to model the relationships among classes, which is quite insensitive to instance level noise. \cite{Weinberger08large} proposed learning mapping functions to embed input vectors and classes into a low dimensional common space based on class taxonomies. \cite{bengio2010label} proposed a label embedding tree method for large multi-class tasks, which also embeds class labels in a low dimensional space. \cite{hwang2013analogy} proposed an analogy-preserving semantic embedding method for multi-class classification. Later \cite{hwang2014unified} proposed a unified semantic embedding method to incorporate different semantic information into learning. Recently \cite{norouziMBSSFCD14} proposed a semantic embedding method for zero-shot learning to embed an unseen class as a convex combination of seen classes with heuristic weights. \cite{DBLP:journals/corr/HammB15} proposed a semantic ranking representation based on semantic similarity to aggregate semantic information from multiple heterogeneous sources. Our embedding is to represent each class as a mixture of seen classes in both domains.

\begin{table}[t]\centering\footnotesize
\begin{tabular}{|p{1.4cm}|p{6cm}|}
  \hline
  Notation & Definition\\
  \hline
  $\mathcal{S}$ ($\mathcal{U}$) & Set of seen (unseen) classes\\
  $|\mathcal{S}|$ & Number of seen classes\\
  $s$ (or $y$) \& $u$ & Indexes for seen and unseen classes\\
  $\Delta^{|\mathcal{S}|}$ & Simplex in $\mathbb{R}^{|\mathcal{S}|}$ dimensional space\\
  $\{\mathbf{c}_y\}$ & Source domain attribute vector $\mathbf{c}_y\in\mathbb{R}^{d_s}$ for class $y$ with $\ell_2$ normalization, \ie $\|\mathbf{c}_y\|=1$. \\
  $\{(\mathbf{x}_i, y_i)\}$ & Training data: $\mathbf{x}_i\in\mathbb{R}^{d_t}$ - target feature, $y_i$ - class\\
  $N (N_y)$ & Number of training samples (for class $y \in {\cal S}$)\\
  $\psi, \pi$ & Source/Target domain feature embedding functions\\    
  $\phi_y$ & Target domain class dependent feature transformation \\
  $(\cdot)_{m,n}$ & The $n$th entry in vector $(\cdot)_m$\\
  \hline
  $\Scale[0.95]{\mathbf{z}_y=\psi(\mathbf{c}_y)}$ & {\bf Learned} source domain embedded histogram $\mathbf{z}_y\in\Delta^{|\mathcal{S}|}$ for class $y$.\\    
  $\mathcal{V}=\{\mathbf{v}_y\}$ & {\bf Learned} target domain reference vector $\mathbf{v}_y\in\mathbb{R}^{d_t}$ for class $y$, one vector per seen class\\    
  $\mathbf{w}$ & {\bf Learned} target domain weight vector\\
  $f(\mathbf{x},y)$ & {\bf Learned} structured scoring function relating the target domain sample $\mathbf{x}$ and class label $y$.\\
  \hline
\end{tabular}
\vspace{1mm}
\caption{\footnotesize Some notation used in our method.}
\vspace{-3mm}
\label{tab:notation}
\end{table}

\section{Zero-Shot Learning and Prediction}\label{sec:method}
\noindent
Our notation is summarized in Table \ref{tab:notation} for future reference. 

\subsection{Overview} \label{ssec:overview}
\noindent
Our method is based on expressing source/target data as a mixture of seen class proportions (see Fig.~\ref{fig:overview}). Using seen class data we learn source and target domain embedding functions, $\psi,\,\pi$ respectively. Our aim is to construct functions that take an arbitrary source vectors $\mathbf{c}$ and target vectors $\mathbf{x}$ as inputs and embed them into $\Delta^{|\S|}$ (histograms). 
Observe that components, $\pi_y(\mathbf{x}),\,\psi_y(\mathbf{c})$ of $\pi(\mathbf{x}),\,\psi(\mathbf{c})$, corresponding to seen class $y \in {\cal S}$, denote the proportion of class $y$ in the instance $\mathbf{x},\,\mathbf{c}$. 
During test-time source domain vectors $\mathbf{c}_u \in {\cal C}$ for all the unseen classes are revealed. We are then presented with an arbitrary target instance $\mathbf{x}$. We predict an unseen label for $\mathbf{x}$ by maximizing the semantic similarity between the histograms. Letting $\mathbf{z}_u=\psi(\mathbf{c}_u)$, then our zero-shot recognition rule is defined as follows:
\begin{align}\label{eqn:f}
u^* = \argmax_{u \in {\cal U}}f(\mathbf{x},u) = \argmax_{u \in {\cal U}} \langle \pi(\mathbf{x}), \mathbf{z}_u \rangle,
\end{align}
where $\langle\cdot,\cdot\rangle$ denotes the inner product of two vectors.


We propose parameterized-optimization problems to learn embedding functions from seen class data. We then optimize these parameters globally using held-out seen class data. We summarize our learning scheme below.

\noindent
{\bf (A)} {\it Source Domain Embedding Function ($\psi$):} Our embedding function is realized by means of a parameterized optimization problem, which is related to sparse coding.

\noindent
{\bf (B)} {\it Target Domain Embedding Function ($\pi$):}
We model $\pi_y(\x)$ as $\langle \mathbf{w}, \phi_y(\mathbf{x})\rangle$. This consists of a constant weight vector $\w$ and a class dependent feature transformation $\phi_y(\mathbf{x})$. We propose a margin-based optimization problem to jointly learn both the weight vector and the feature transformation. Note that our parameterization may yield negative values and may not be normalized, which can be incorporated as additional constraints but we ignore this issue in our optimization objectives.

\noindent
{\bf (C)} {\it Cross Validation:} Our embedding functions are parameter dependent. We choose these parameters by employing a cross validation technique based on holding out seen class data. First, we learn embedding functions (see (A) and (B)) on the remaining (not held-out) seen class data with different values of the predefined parameters. We then jointly optimize parameters of source/target embedding functions to minimize the prediction error on held-out seen classes. In the end we re-train the embedding functions over the entire seen class data.

\noindent
{\bf Salient Aspects of Proposed Method:} 

\noindent
\textbf{(a)} {\it Decomposition:} Our method seeks to decompose source and target domain instances into mixture proportions of seen classes. In contrast much of the existing work can be interpreted as learning cross-domain similarity between source domain attributes and target feature components. 

\noindent
\textbf{(b)} {\it Class Dependent Feature Transformation $\pi_y(\x)$:} The decomposition perspective necessitates fundamentally new design choices. For instance, $\pi_y(\x)$, the component corresponding to class $y$ must be dependent on $y$, which implies that we must choose a class dependent feature transform $\phi_y(\x)$ because $\w$ is a constant vector and agnostic to class.

\noindent
\textbf{(c)} {\it Joint Optimization and Generalization to Unseen Classes:} Our method jointly optimizes parameters of the embedding functions to best align source and target domain histograms for held-out seen classes, thus providing a basis for generalizing to unseen classes. Even for fixed parameters, embedding functions $\psi,\,\pi$ are {\bf nonlinear} maps and since the parameters are jointly optimized our learned scoring function $f(\mathbf{x},y)$ couples seen source and target domain together in a rather complex way. So we cannot reduce $f(\cdot,\cdot)$ to a linear or bilinear setting as in \cite{Akata2015}.

\subsection{Intuitive Justification of Proposed Method}\label{ssec:justification}
\noindent
Recall that our method is based on viewing unseen source and target instances as a histogram of seen classes proportions. Fig. \ref{fig:overview} suggests that a target instance can be viewed as arising from a mixture of seen classes with mixture components dependent on the location of the instance. More precisely, letting $P$ and $P_y$ be the unseen and seen class-conditional target feature distributions respectively, we can a priori approximate $P$ as a mixture of the $P_y$'s, \ie $P = \sum_{y\in {\cal S}} \bar \pi_y P_y + P_{\mbox{error}}$ (see \cite{blanchard2014decontamination} for various approaches in this context), where $\bar \pi_y$ denotes the mixture weight for class $y$. Analogously, we can also decompose source domain data as a mixture of source domain seen classes. This leads us to associate mixture proportion vector $\mathbf{z}_u$ with unseen class $u$, and represent attribute vector $\mathbf{c}_u$ as $\mathbf{c}_u \approx \sum_{y \in {\cal S}} z_{u,y} \mathbf{c}_y$, with $\mathbf{z}_u = (z_{u,y})_{y \in {\cal S}} \in \Delta^{|\S|}$. 

\noindent
{\bf Key Postulate:} The target domain instance, $\x$, must have on average a similar mixture pattern as the source domain pattern if they both correspond to the same unseen label, $u \in {\cal U}$, namely, on average $\pi(\x)$ is equal to $\mathbf{z}_u$.
%
%
%

This postulate is essentially Eq.~\ref{eqn:f}. This postulate also motivates our margin-based approach for learning $\w$. 
Note that since we only have a single source domain vector for each class, a natural constraint is to require that the empirical mean of the mixture corresponding to each example per class in target domain aligns well with the source domain mixture. This is empirically consistent with our postulate. Letting $y, y'$ be seen class labels with $y \neq y'$ and $\bar{\boldsymbol{\pi}}_y$ denote the average mixture for class $y$ in target domain, our requirement is to guarantee that
\begin{align}\label{eqn:alignment}
& \langle \bar{\boldsymbol{\pi}}_y, \mathbf{z}_y \rangle \geq \langle \bar{\boldsymbol{\pi}}_y, \mathbf{z}_{y'} \rangle \\
&\Leftrightarrow\sum_{s \in {\cal S}} \left\langle \mathbf{w}, \underbrace{{1 \over N_s} \sum_{i=1}^N \mathbb{I}_{\{y_i=s\}}\phi_s(\mathbf{x}_i)}_{\mbox{\footnotesize Emp. Mean Embedding}} \right\rangle  \Big(z_{y,s} - z_{y',s}\Big)\geq 0, \nonumber
\end{align}
where $\mathbb{I}_{\{\cdot\}}$ denotes a binary indicator function returning 1 if the condition holds, otherwise 0.
Note that the empirical mean embedding corresponds to a kernel empirical mean embedding \cite{smola2007hilbert} if $\phi_s$ is a valid (characteristic) RKHS kernel, but we do not pursue this point further in this paper.
%
Nevertheless this alignment constraint is generally insufficient, because it does not capture the shape of the underlying sample distribution. We augment misclassification constraints for each seen sample in SVMs to account for shape.

\subsection{Source Domain Embedding}\label{ssec:sdse}
Recall from Fig.~\ref{fig:overview} and {\bf (B)} in Sec.~\ref{ssec:overview} that our embedding aims to map source domain attribute vectors $\mathbf{c}$ to histograms of seen class proportions, \ie $\psi: \mathbb{R}^{d_s} \rightarrow \Delta^{|{\cal S}|}$. We propose a parameterized optimization problem inspired by sparse coding as follows, given a source domain vector $\mathbf{c}$: 
\begin{equation}\label{eqn:simple}
\psi(\mathbf{c})=
\argmin_{\boldsymbol{\alpha}\in \Delta^{|{\cal S}|}} \left\{\frac{\gamma}{2}\|\boldsymbol{\alpha}\|^2+\frac{1}{2}\|\mathbf{c}-\sum_{y\in\mathcal{S}}\mathbf{c}_y \alpha_y\|^2\right\},
\end{equation}
where $\gamma\geq 0$ is a predefined regularization parameter, $\|\cdot\|$ denotes the $\ell_2$ norm of a vector, and $\boldsymbol{\alpha} = (\alpha_y)_{y \in {\cal S}}$ describes contributions of different seen classes. Note that even though $\mathbf{c}$ may not be on the simplex, the embeddings $\psi(\mathbf{c})$ are always. 
Note that the embedding $\psi$ is in general a nonlinear function. Indeed on account of simplex constraint small values in $\boldsymbol{\alpha}$ vector are zeroed out (\ie ``water-filling'').

\begin{figure}[t]
\begin{minipage}[b]{0.49\linewidth}
 \begin{center}
 \centerline{\includegraphics[width=.95\columnwidth]{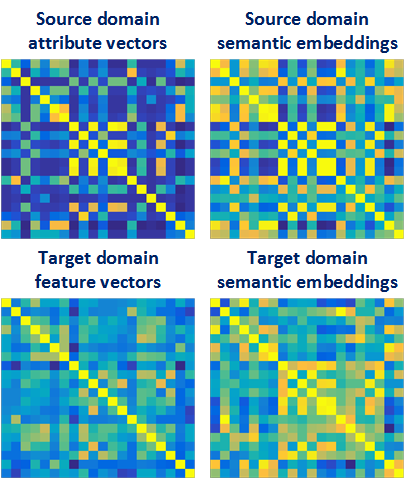}}
 \centerline{\footnotesize{(a) Seen classes in training}}
 \end{center}
\end{minipage}
\begin{minipage}[b]{0.49\linewidth}
\begin{center}
\centerline{\includegraphics[width=.95\columnwidth]{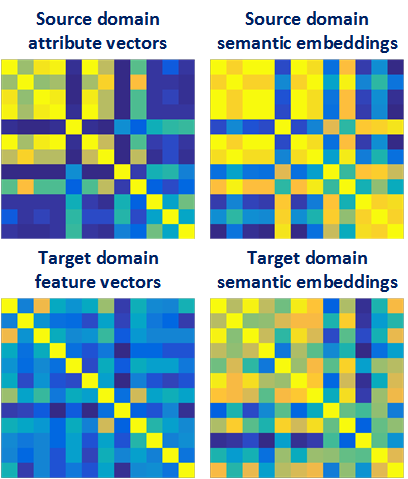}}
\centerline{\footnotesize{(b) Unseen classes in testing}}
\end{center} 
\end{minipage}
\vspace{-3mm}
\caption{\footnotesize{Cosine similarity matrices among (a) seen and (b) unseen classes on aPascal \& aYahoo \cite{farhadi2009attribute} dataset. Brighter color depicts larger values. The type of data used to compute the matrix is shown above the corresponding matrix. Observe that in training/testing our source/target domain embedding preserves the inter-class relationships originally defined by the source domain attribute vectors. This also indicates that our target domain embeddings manage to align well the target domain distributions with the source domain attribute vectors. 
}}\label{fig:source_domain_embedding}
%
\vspace{-3mm}
\end{figure}

To solve Eq. \ref{eqn:simple}, we use quadratic programming. For large-scale cases, we adopt efficient proximal gradient descent methods. 
Note that there are many alternate ways of embedding such as similarity rescaling, subspace clustering \cite{peng2012constructing}, sparse learning \cite{elhamifar2013sparse}, and low rank representation \cite{liu2013robust}, as long as the embedding is on the simplex. We tried these different methods with the simplex constraint to learn the embeddings, and our current solution in Eq. \ref{eqn:simple} works best. We believe that it is probably because the goal in these other methods is subspace clustering, while our goal is to find a noise resilient embedding which has good generalization to unseen class classification. 

We optimize the parameter, $\gamma$, globally by cross validation. Once the $\gamma$ parameter is identified, all of the seen classes are used in our embedding function.
Note that when $\gamma=0$ or small, $\psi(\mathbf{c}_y)$ will be a coordinate vector, which essentially amounts to coding for multi-class classification but is not useful for unseen class generalization. Conceptually, because we learn tuning parameters to predict well on held-out seen classes, $\gamma$ is in general not close to zero. 
We demonstrate class affinity matrices before and after embedding for both seen and unseen classes in Fig. \ref{fig:source_domain_embedding}. Here $\gamma=10$ is obtained by cross validation. We see that in both training and testing source domain embeddings preserve the affinities among classes in the attribute space.

During test-time when unseen class attribute vectors $\mathbf{c}_u$ are revealed, we obtain $\mathbf{z}_u$ as the embeddings using Eq. \ref{eqn:simple} with the learned $\gamma$.

\subsection{Target Domain Embedding}\label{ssec:tdse}

In this paper we define our target domain class dependent mapping function $\phi_y$ based on (1) intersection function (INT) \cite{maji2008classification}, or (2) rectified linear unit (ReLU) \cite{nair2010rectified}. That is,
\begin{align}
\mbox{INT:} & \quad \phi_y(\mathbf{x})=\min(\mathbf{x}, \mathbf{v}_y), \label{eqn:int}\\
\mbox{ReLU:} & \quad \phi_y(\mathbf{x})=\max(\mathbf{0}, \mathbf{x}-\mathbf{v}_y), \label{eqn:relu}
\end{align}
where $\min$ and $\max$ are the entry-wise operators. Note that intersection function captures the data patterns in $\mathbf{x}$ below the thresholds in each $\mathbf{v}_y$, while ReLU captures the data patterns above the thresholds. In this sense, the features generated from these two functions are complementary. This is the reason that we choose the two functions to demonstrate the robustness of our method.

Based on Eq. \ref{eqn:f} and \ref{eqn:alignment} in Section \ref{ssec:overview}, we define the following structured scoring function $f(\mathbf{x},y)$ as follows:
\begin{align}\label{eqn:Psi}
f(\mathbf{x},y)=\sum_{s\in\mathcal{S}}\left\langle\mathbf{w}, \phi_s(\mathbf{x})\right\rangle z_{y,s}.
\end{align}
In test-time for target instance $\mathbf{x}$, we can compute $f(\mathbf{x},u)$ for an arbitrary unseen label $u$ because the source attribute vector is revealed for $u$. Note that $f$ is highly non-convex, and it cannot reduce to bilinear functions used in existing works such as \cite{akata2013label,Akata2015}.

\subsubsection{Max-Margin Formulation}
Based on Eq. \ref{eqn:Psi}, we propose the following parameterized learning formulation for zero-shot learning as follows, which learns the embedding function $\pi$, and thus $f$:
\begin{align}\label{eqn:sl}
&\min_{\mathcal{V}, \mathbf{w}, \boldsymbol{\xi}, \boldsymbol{\epsilon}}\frac{1}{2}\|\mathbf{w}\|^2+\frac{\lambda_1}{2}\sum_{\mathbf{v}\in\mathcal{V}}\|\mathbf{v}\|^2+\lambda_2\sum_{y,s}\epsilon_{ys}+\lambda_3\sum_{i,y}\xi_{iy}\\
&\mbox{s.t.} \; \forall i\in\{1,\cdots,N\}, \forall y\in\mathcal{S}, \forall s\in\mathcal{S}, \nonumber\\
&\sum_{i=1}^N {\mathbb{I}_{\{y_i=y\}} \over N_y}\Big[f(\mathbf{x}_i,y)-f(\mathbf{x}_i,s)\Big]\geq\Delta(y,s)-\epsilon_{ys}, \label{eqn:mean}\\
&  f(\mathbf{x}_i,y_i)-f(\mathbf{x}_i,y)\geq\Delta(y_i,y)-\xi_{iy}, \label{eqn:instance}\\
& \epsilon_{ys}\geq0, \xi_{iy}\geq0, \forall \mathbf{v}\in\mathcal{V}, \mathbf{v}\geq\mathbf{0},\nonumber
\end{align}
where $\Delta(\cdot, \cdot)$ denotes a structural loss between the ground-truth class and the predicted class, $\lambda_1\geq0$, $\lambda_2\geq0$, and $\lambda_3\geq0$ are the predefined regularization parameters, $\boldsymbol{\xi}=\{\xi_{iy}\}$ and $\boldsymbol{\epsilon}=\{\epsilon_{ys}\}$ are slack variables, and $\mathbf{0}$ is a vector of 0's. In this paper, we define $\Delta(y_i,y)=1-\mathbf{c}_{y_i}^T\mathbf{c}_{y}$ and $\Delta(y,s)=1-\mathbf{c}_{y}^T\mathbf{c}_{s}$, respectively. Note that in learning we only access and utilize the data from seen classes.

\begin{figure}[t]
\begin{minipage}[b]{0.325\linewidth}
 \begin{center}
 \centerline{\includegraphics[width=.9\columnwidth]{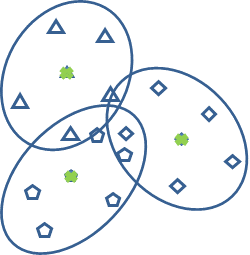}}
 \centerline{\footnotesize{(a) Distr. alignment}}
 \end{center}
\end{minipage}
\begin{minipage}[b]{0.325\linewidth}
\begin{center}
\centerline{\includegraphics[width=.9\columnwidth]{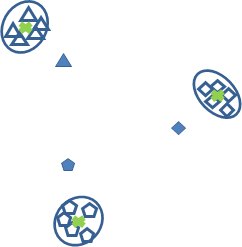}}
 \centerline{\footnotesize{(b) Inst. classification}}
\end{center} 
\end{minipage}
\begin{minipage}[b]{0.325\linewidth}
\begin{center}
\centerline{\includegraphics[width=.9\columnwidth]{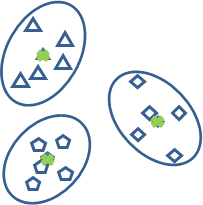}}
 \centerline{\footnotesize{(c) Our method}}
\end{center} 
\end{minipage}
\vspace{-7mm}
\caption{\footnotesize{Illustration of three different constraints for learning the target domain semantic embedding function. Different shapes denote differnt classes, fill-in shapes denote the source domain embeddings, and green crosses denote the empirical means of target domain data embeddings. Our method takes into account the zero-shot learning based on both distribution alignment and instance classification.}}\label{fig:cluster}
\vspace{-3mm}
\end{figure}

In fact, Eq. \ref{eqn:mean} measures the alignment loss for each seen class distribution, and Eq. \ref{eqn:instance} measures the classification loss for each target domain training instance, respectively, which correspond to the discussion in Sec. \ref{ssec:justification}. On one hand, if we only care about the alignment condition, it is likely that there may be many misclassified training data samples (\ie loose shape) as illustrated in Fig. \ref{fig:cluster}(a). On the other hand, conventional classification methods only consider separating data instances with tight shape, but are unable to align distributions due to lack of such constraint in training (see Fig. \ref{fig:cluster}(b)). 
By introducing these two constraints into Eq.~\ref{eqn:sl}, we are able to learn the target domain embedding function as well as the scoring function to produce the clusters which are well aligned and separated, as illustrated in Fig. \ref{fig:cluster}(c). 

Similarly, we learn the predefined parameters $\lambda_1,\,\lambda_2,\,\lambda_3$ through a cross validation step that optimizes the prediction for held-out seen classes. Then once the parameters are determined we re-learn the classifier on all of the seen data. Fig. \ref{fig:source_domain_embedding} depicts class affinity matrices before and after target domain semantic embedding on real data. Our method manages to align source/target domain data distributions.

\subsubsection{Alternating Optimization Scheme}  
To solve Eq. \ref{eqn:sl}, we propose the following alternating optimization algorithm, as seen in Alg. \ref{alg:1}.
\begin{algorithm}[ht]\footnotesize
\SetAlgoLined
\SetKwInOut{Input}{Input}\SetKwInOut{Output}{Output}
\Input{$\{\mathbf{x}_i,y_i\}$, $\{\mathbf{c}_y\}_{y\in\mathcal{S}}$, $\{\mathbf{z}_y\}_{y\in\mathcal{S}}$, $\lambda_1$, $\lambda_2$, $\lambda_3$, learning rate $\eta\geq0$}
Initialize $\boldsymbol{\nu}^{(0)}$ with feature means of seen classes in target domain;\\
\For{$t=0$ \emph{\KwTo} $\tau$}
{ 
$(\mathbf{w}, \boldsymbol{\epsilon}, \boldsymbol{\xi})\leftarrow \mbox{linearSVM\_solver}(\{\mathbf{x}_i,y_i\},\boldsymbol{\nu}^{(t)},\lambda_2, \lambda_3)$;\\
$\boldsymbol{\nu}^{(t+1)}\leftarrow\max\{\mathbf{0}, \boldsymbol{\nu}^{(t)}-\eta\nabla h(\boldsymbol{\nu}^{(t)})\}$;\\
Check monotonic decreasing condition on the objective in Eq. \ref{eqn:sl};
}
\Output{$\mathbf{w}, \boldsymbol{\nu}$}
\caption{Learning Embedding Functions}\label{alg:1}
\end{algorithm}

\noindent
{\bf (\rmnum{1}) Learning $\mathbf{w}$ by fixing $\mathcal{V}$:} In this step, we can collect all the constraints in Eq. \ref{eqn:mean} and Eq. \ref{eqn:instance} by plugging in $\{(\mathbf{x}_i, y_i)\}, \mathcal{V}, \{\mathbf{c}_y\}_{y\in\mathcal{S}}$, and then solve a linear SVM to learn $\mathbf{w}, \boldsymbol{\epsilon}, \boldsymbol{\xi}$, respectively.

\noindent
{\bf (\rmnum{2}) Learning $\mathcal{V}$ by fixing $\mathbf{w}$ using Concave-Convex procedure (CCCP) \cite{yuille2003concave}:} Note that the constraints in Eq. \ref{eqn:mean} and Eq. \ref{eqn:instance} consist of difference-of-convex (DoC) functions. To see this, we can rewrite $f(\mathbf{x}_i,y)-f(\mathbf{x}_i,y_i)$ as a summation of convex and concave functions as follows:
\begin{equation}\label{eqn:cccp}
f(\mathbf{x}_i,y)-f(\mathbf{x}_i,y_i) =\sum_{m,s}w_m(z_{y,n}-z_{y_i,n})\phi_{s,m}(\mathbf{x}_i),
\end{equation}
where $w_m$ and $\phi_{s,m}(\cdot)$ denote the $m$th entries in vectors $\mathbf{w}$ and $\phi_s(\cdot)$, respectively. Let $\boldsymbol{\nu}\in\mathbb{R}^{d_t|\mathcal{S}|}$ be a vector concatenation of all $\mathbf{v}$'s, $g_1(\boldsymbol{\nu})\stackrel{\Delta}{=}g_1(\mathbf{x}_i,y,\boldsymbol{\nu})$ and $g_2(\boldsymbol{\nu})\stackrel{\Delta}{=}g_2(\mathbf{x}_i,y,\boldsymbol{\nu})$ denote the summations of all the convex and all the concave terms in Eq.~\ref{eqn:cccp}, respectively. Then we have $f(\mathbf{x}_i,y)-f(\mathbf{x}_i,y_i)=g_1(\boldsymbol{\nu})-(-g_2(\boldsymbol{\nu}))$, \ie DoC functions. Using CCCP we can relax the constraint in Eq. \ref{eqn:instance} as $\xi_{iy}\geq\Delta(y_i,y)+g_1(\boldsymbol{\nu})+g_2(\boldsymbol{\nu}^{(t)})+\nabla g_2(\boldsymbol{\nu}^{(t)})^T(\boldsymbol{\nu}-\boldsymbol{\nu}^{(t)})$,
where $\boldsymbol{\nu}^{(t)}$ denotes the solution for $\boldsymbol{\nu}$ in iteration $t$, and $\nabla$ denotes the subgradient operator. Similarly we can perform CCCP to relax the constraint in Eq. \ref{eqn:mean}. Letting $h(\boldsymbol{\nu})$ denote the minimization problem in Eq. \ref{eqn:sl}, \ref{eqn:mean}, and \ref{eqn:instance}, using CCCP we can further write down the subgradient $\nabla h(\boldsymbol{\nu}^{(t)})$ in iteration $t+1$ as follows:
\begin{align}\label{eqn:h}
& \hspace{-5mm}\nabla h(\boldsymbol{\nu}^{(t)})=\lambda_1\boldsymbol{\nu}^{(t)}\nonumber\\
& +\lambda_2\sum_{y,s,i}\mathbb{I}_{\{\epsilon_{ys}>0,y_i=y\}}\left[\nabla g_1(\boldsymbol{\nu}^{(t)})+\nabla g_2(\boldsymbol{\nu}^{(t)})\right]\nonumber\\
& +\lambda_3\sum_{y_i,y}\mathbb{I}_{\{\xi_{iy}>0\}}\left[\nabla g_1(\boldsymbol{\nu}^{(t)})+\nabla g_2(\boldsymbol{\nu}^{(t)})\right].
\end{align}
Then we use subgradient descent to update $\boldsymbol{\nu}$, equivalently learning $\mathcal{V}$. With simple algebra, we can show that the $m$th entry for class $n$ in $\nabla g_1(\boldsymbol{\nu}^{(t)})+\nabla g_2(\boldsymbol{\nu}^{(t)})$ is equivalent to the $m$th entry in $\left.\frac{\partial f(\mathbf{x}_i,y)}{\partial \mathbf{v}_s}\right|_{\boldsymbol{\nu}^{(t)}}-\left.\frac{\partial f(\mathbf{x}_i,y_i)}{\partial \mathbf{v}_s}\right|_{\boldsymbol{\nu}^{(t)}}$. In order to guarantee the monotonic decrease of the objective in Eq. \ref{eqn:sl}, we add an extra checking step in each iteration. 

\subsection{Cross Validation on Seen Class Data}
The scoring function in Eq.~\ref{eqn:Psi} is obtained by solving Eq.~\ref{eqn:simple} and ~\ref{eqn:sl}, which in turn depend on parameters $\boldsymbol{\theta} = (\gamma, \lambda_1, \lambda_2, \lambda_3)$. We propose learning these parameters by means of cross validation using held-out seen class data. Specifically, define ${\cal S}_{\ell} \subset {\cal S}$ and the held-out set ${\cal S}_h = {\cal S} \backslash {\cal S}_{\ell}$. We learn a collection of embedding functions for source and target domains using Eq.~\ref{eqn:simple} and ~\ref{eqn:sl} over a range of parameters $\boldsymbol{\theta}$ suitably discretized in 4D space. For each parameter choice $\boldsymbol{\theta}$ we obtain a scoring function, which depends on training subset as well as the parameter choice. We then compute the prediction error, namely, the number of times that a held-out target domain sample is misclassified for this parameter choice. We repeat this procedure for different randomly selected subsets ${\cal S}_{\ell}$ and choose parameters with the minimum average prediction error.  
%
Once these parameters are obtained we then plug it back into Eq. \ref{eqn:simple} and \ref{eqn:sl}, and re-learn the scoring function using all the seen classes.

\section{Experiments}\label{sec:exp}
We test our method on five benchmark image datasets for zero-shot recognition, \ie CIFAR-10 \cite{citeulike:7491128}, aPascal \& aYahoo (aP\&Y) \cite{farhadi2009attribute}, Animals with Attributes (AwA) \cite{citeulike:7491128}, Caltech-UCSD Birds-200-2011 (CUB-200-2011) \cite{WahCUB_200_2011}, and SUN Attribute \cite{patterson2014sun}. For all the datasets, we utilize MatConvNet \cite{arXiv:1412.4564} with the ``imagenet-vgg-verydeep-19'' pretrained model \cite{simonyan2014very} to extract a 4096-dim CNN feature vector (\ie the top layer hidden unit activations of the network) for each image (or bounding box). Verydeep features work well since they lead to good class separation, which is required for our class dependent transform (see Fig. \ref{fig:awa-feat}). Similar CNN features were used in previous work \cite{Akata2015} for zero-shot learning. We denote the two variants of our general method as {\em SSE-INT} and {\em SSE-ReLU}, respectively. Note that in terms of experimental settings, the main difference between our method and the competitors is the features. We report the top-1 recognition accuracy averaged over 3 trials.

We set $\gamma,\lambda_2,\lambda_3\in\{0, 10^{-3}, 10^{-2}, 10^{-1}, 1, 10, 10^{2}\}$ in Eq. \ref{eqn:simple} and \ref{eqn:sl} for cross validation. In each iteration, we randomly choose two seen classes for validation, and fix $\boldsymbol{\nu}$ in Alg. \ref{alg:1} to its initialization for speeding up computation. For $\lambda_1$, we simply set it to a small number $10^{-4}$ because it is much less important than the others for recognition.


\subsection{CIFAR-10}

\begin{table*}[t]\footnotesize
\centering
\caption{\footnotesize{Zero-shot recognition accuracy comparison (\%, mean$\pm$standard deviation) on CIFAR-10. The compared numbers are best estimated from Fig. 3 in \cite{socher2013zero}. Notice that all the methods here utilize deep features to represent images in target domain.}}\label{tab:cifar-10}\vspace{1mm}
\begin{tabular}{|l|lllll|l|}
\hline
 Method & cat-dog & plane-auto & auto-deer & deer-ship & cat-truck & Average \\
 \hline\hline  
 Socher \etal \cite{socher2013zero} (50 words) & 50 & 65 & 76 & 83 & 90 & 72.8 \\
 \hline\hline
 SSE-INT (50 words) & \textbf{\em 59.00$\pm$0.57} & 91.62$\pm$0.19 & 97.95$\pm$0.13 & 95.73$\pm$0.08 & 97.20$\pm$0.05 & \textbf{\em 88.30} \\
 SSE-ReLU (50 words) & 58.78$\pm$1.60 & 91.33$\pm$0.53 & 97.33$\pm$0.28 & 95.37$\pm$0.29 & \textbf{\em 97.32$\pm$0.12} & 88.03 \\
 \hline\hline
 SSE-INT (25-dim binary vectors) & 48.47$\pm$0.08 & \textbf{\em 93.93$\pm$0.59} & \textbf{\em 99.07$\pm$0.18} & \textbf{\em 96.03$\pm$0.03} & 96.92$\pm$0.14 & 86.88 \\
 SSE-ReLU (25-dim binary vectors) & 48.52$\pm$0.13 & 93.68$\pm$0.73 & 98.48$\pm$0.15 & 95.32$\pm$0.25 & 96.43$\pm$0.06 & 86.49 \\
 \hline
\end{tabular}
\end{table*}

\begin{table*}[t]\footnotesize
\begin{minipage}{\textwidth}
\centering
\caption{\footnotesize{Zero-shot recognition accuracy comparison (\%) on aP\&Y, AwA, CUB-200-2011, and SUN Attribute, respectively, in the form of mean$\pm$standard deviation. Here except our results, the rest numbers are cited from their original papers. Note that some experimental settings may differ from ours.}}\label{tab:apy}\vspace{1mm}
\begin{tabular}{|l|lllll|}
\hline
Feature & Method & aPascal \& aYahoo & Animals with Attributes & CUB-200-2011 & SUN Attribute \\
\hline\hline
\multirow{11}{*}{Non-CNN} &
Farhadi \etal \cite{farhadi2009attribute} & 32.5 & & & \\
& Mahajan \etal \cite{mahajan2011joint} & 37.93 & & & \\
& Wang and Ji \cite{wang2013unified} & 45.05 & 42.78 & & \\
& Rohrbach \etal \cite{conf/nips/RohrbachES13} & & 42.7 & & \\
& Yu \etal \cite{yu2013designing} & & 48.30 & & \\
& Akata \etal \cite{akata2013label} & & 43.5 & 18.0 & \\
& Fu \etal \cite{embedding2014ECCV} & & 47.1 & & \\
& Mensink \etal \cite{mensink2014costa} & & & 14.4 & \\
& Lampert \etal \cite{10.1109/TPAMI.2013.140} & 19.1 & 40.5 & & 52.50 \\
& Jayaraman and Grauman \cite{jayaraman2014unreliable} & 26.02$\pm$0.05 & 43.01$\pm$0.07 & & 56.18$\pm$0.27 \\
& Romera-Paredes and Torr \cite{Romera-Paredes2015} & 27.27$\pm$1.62 & 49.30$\pm$0.21 & & 65.75$\pm$0.51 \\
\hline\hline
AlexNet &
Akata \etal \cite{Akata2015}\footnote{The results listed here are the ones with 4096-dim CNN features and the continuous attribute vectors provided in the datasets for fair comparison. } & & 61.9 & \textbf{\em 40.3} & \\
\hline\hline
\multirow{4}{*}{vgg-verydeep-19} & Lampert \etal \cite{10.1109/TPAMI.2013.140} & 38.16 & 57.23 & & 72.00 \\
& Romera-Paredes and Torr \cite{Romera-Paredes2015} & 24.22$\pm$2.89 & 75.32$\pm$2.28 & & 82.10$\pm$0.32 \\
& SSE-INT & 44.15$\pm$0.34 & 71.52$\pm$0.79 & 30.19$\pm$0.59 & 82.17$\pm$0.76\\
& SSE-ReLU & \textbf{\em 46.23$\pm$0.53} & \textbf{\em 76.33$\pm$0.83} & 30.41$\pm$0.20 & \textbf{\em 82.50$\pm$1.32}\\
\hline
\end{tabular}
\end{minipage}\vspace{-3mm}
\end{table*}

\begin{figure}[t]
\begin{minipage}[b]{0.495\columnwidth}
 \begin{center}
 \centerline{\includegraphics[width=0.9\columnwidth]{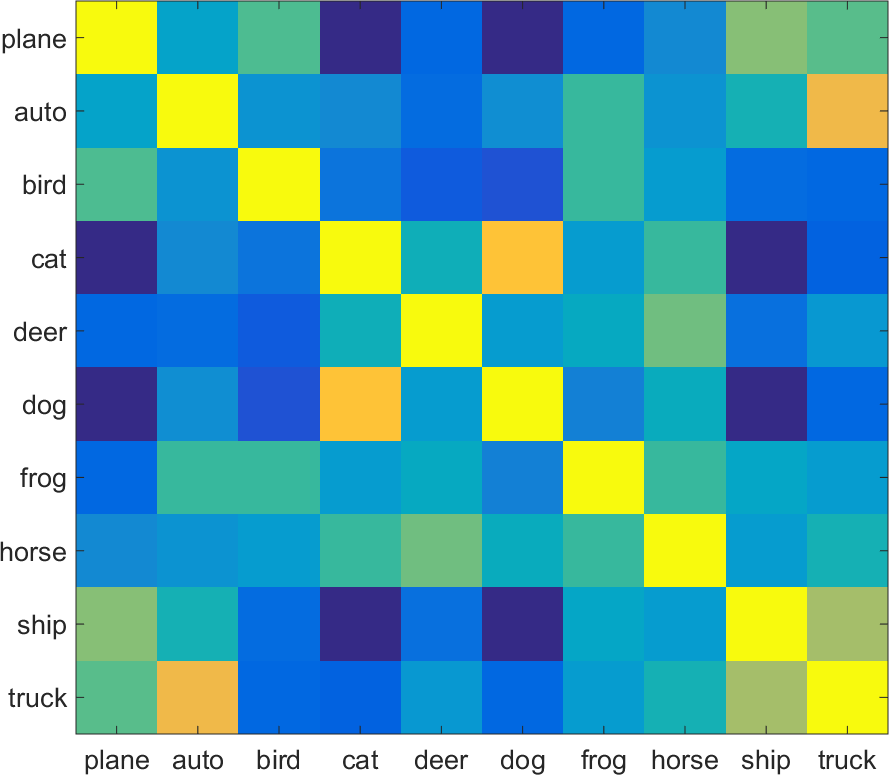}}
 \centerline{\footnotesize{(a) Cosine similarity matrix}}
 \end{center}
\end{minipage}
\begin{minipage}[b]{0.495\columnwidth}
 \begin{center}
 \centerline{\includegraphics[width=0.9\columnwidth]{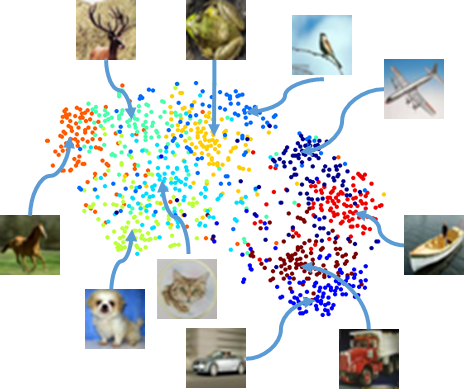}}
 \centerline{\footnotesize{(b) CNN features}}
 \end{center}
\end{minipage}
\begin{minipage}[b]{0.42\columnwidth}
\begin{center}
\centerline{\includegraphics[width=0.9\columnwidth]{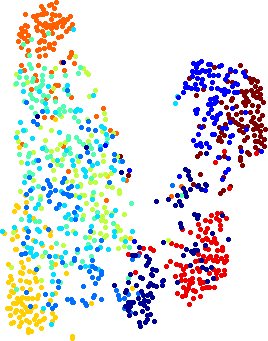}}
 \centerline{\footnotesize{(c) SSE embeddings (auto-deer)}}
\end{center} 
\end{minipage}
\begin{minipage}[b]{0.57\columnwidth}
\begin{center}
\centerline{\includegraphics[width=0.9\columnwidth]{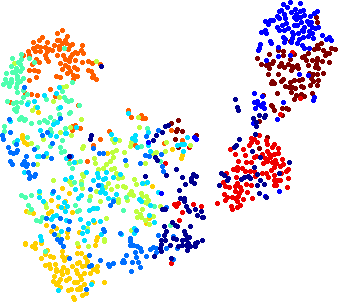}}
 \centerline{\footnotesize{(d) SSE embeddings (cat-dog)}}
\end{center} 
\end{minipage}
\vspace{-7mm}
\caption{\footnotesize{\textbf{(a)} Class affinities for the 10 classes using source domain binary attribute vectors. \textbf{(b-d)} t-SNE visualization of different features with 25 attributes, where 100 samples per class in the test set are selected randomly and the same color denotes the same class. (b) shows the 4096-dim original target domain CNN features. (c) and (d) show the 8-dim learned SSE features by SSE-INT and tested on auto-deer and cat-dog, respectively. The embeddings produced by SSE-ReLU have similar patterns.}}\label{fig:tsne}
\vspace{-3mm}
\end{figure}

This dataset consists of 60000 color images with resolution of $32\times 32$ pixels (50000 for training and 10000 for testing) from 10 classes. \cite{socher2013zero} enriched it with 25 binary attributes and 50-dim semantic word vectors with real numbers for each class. We follow the settings in \cite{socher2013zero}. Precisely, we take cat-dog, plane-auto, auto-deer, deer-ship, and cat-truck as test categories for zero-shot recognition, respectively, and use the rest 8 classes as seen class data. Our training and testing is performed on the split of training and test data provided in the dataset, respectively.

We first summarize the accuracy of \cite{socher2013zero} and our method in Table \ref{tab:cifar-10}. Clearly our method outperforms \cite{socher2013zero} significantly, and SSE-INT and SSE-ReLU perform similarly. We observe that for cat-dog our method performs similarly as \cite{socher2013zero}, while for others our method can easily achieve very high accuracy. We show the class affinity matrix in Fig. \ref{fig:tsne}(a) using the binary attribute vectors, and it turns out that cat and dog have a very high similarity. Similarly the word vectors between cat and dog provide more discrimination than attribute vectors but still much less than others.

To better understand our SSE learning method, we visualize the target domain CNN features as well as the learned SSE features using t-SNE \cite{van2008visualizing} in Fig. \ref{fig:tsne}(b-d). Due to different seen classes, the learned functions and embeddings for Fig. \ref{fig:tsne}(c) and Fig. \ref{fig:tsne}(d) are different. In Fig. \ref{fig:tsne}(b), CNN features seem to form clusters for different classes with some overlaps, and there is a small gap between ``animals'' and ``artifacts''. In contrast, our SSE features are guided by source domain attribute vectors, and indeed preserve the affinities between classes in the attribute space. In other words, our learning algorithm manages to align the target domain distributions with their corresponding source domain embeddings in SSE space, as well as discriminating each target domain instance from wrong classes. As we see, the gaps between animals and artifacts are much clearer in Fig. \ref{fig:tsne}(c) and Fig. \ref{fig:tsne}(d) than that in Fig. \ref{fig:tsne}(b). For cat and dog, however, there is still a large overlap in SSE space, leading to poor recognition. The overall sample distributions in Fig. \ref{fig:tsne}(c) and Fig. \ref{fig:tsne}(d) are similar, because they both preserve the same class affinities.


\subsection{Other Benchmark Comparison}

\begin{figure}[t]
\begin{minipage}[b]{0.495\columnwidth}
 \begin{center}
 \centerline{\includegraphics[width=.9\columnwidth]{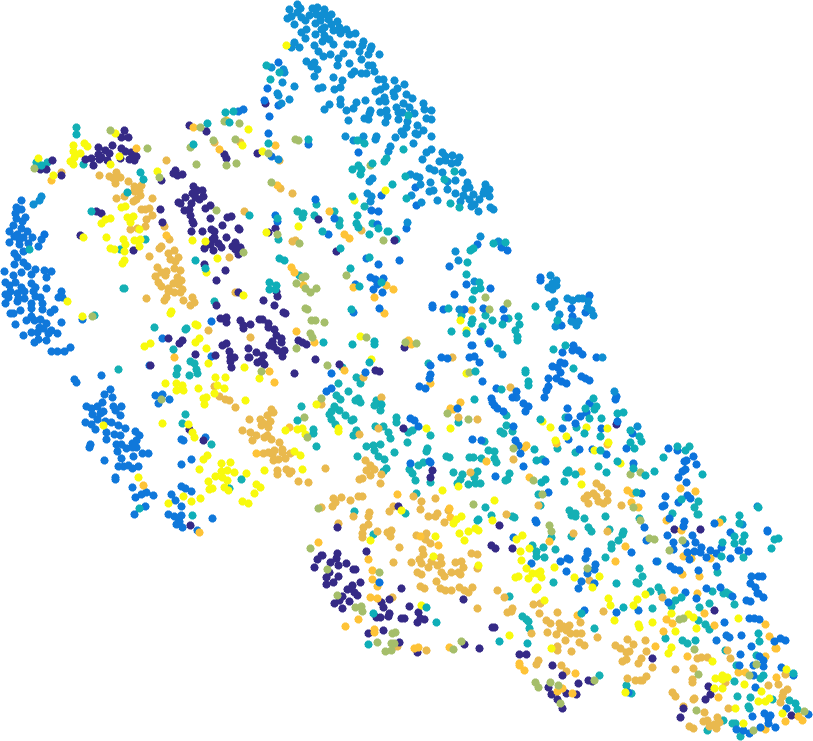}}
 \centerline{\footnotesize{(a) decaf}}
 \end{center}
\end{minipage}
\begin{minipage}[b]{0.495\columnwidth}
 \begin{center}
 \centerline{\includegraphics[width=.9\columnwidth]{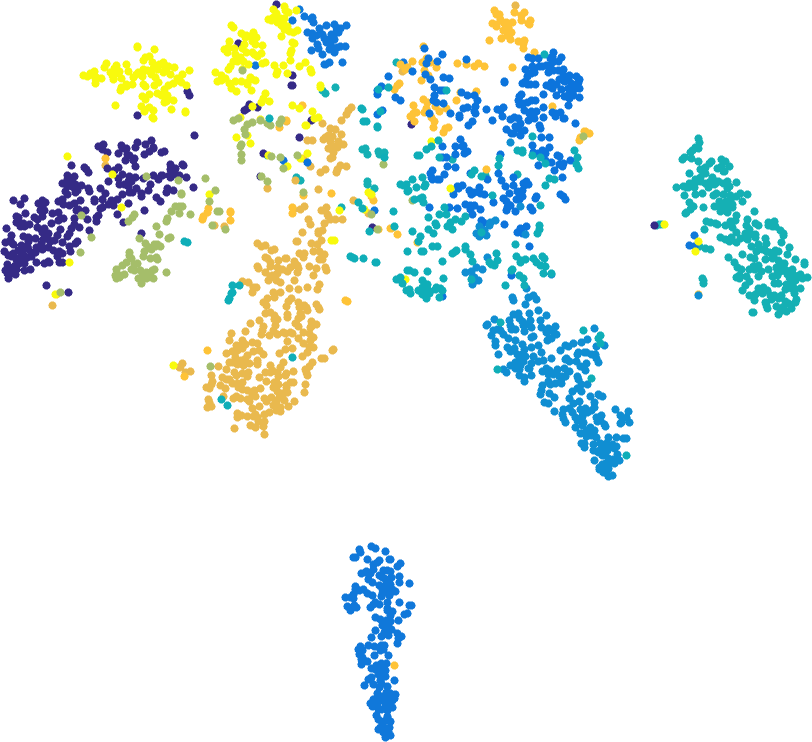}}
 \centerline{\footnotesize{(b) verydeep-19}}
 \end{center}
\end{minipage}
\vspace{-7mm}
\caption{\footnotesize{t-SNE visualization comparison of SSE distributions using the two CNN features on AwA testing data. Our method works well if there is good separation for classes and verydeep features are particularly useful.}}\label{fig:awa-feat}
\vspace{-3mm}
\end{figure}

\begin{figure*}[t]
\begin{minipage}[b]{0.121\linewidth}
 \begin{center}
 \centerline{\includegraphics[width=.9\columnwidth]{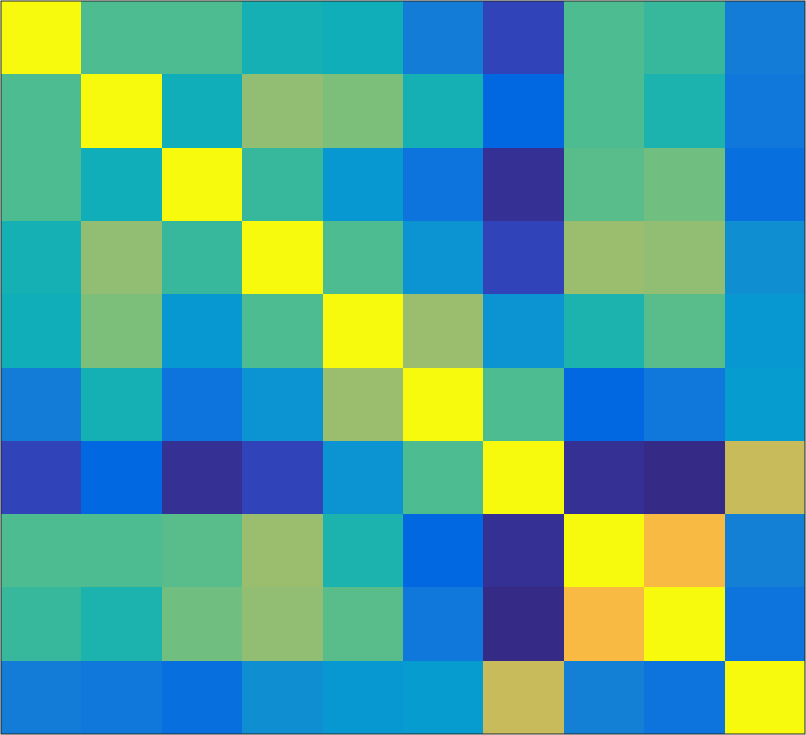}}
 \centerline{\footnotesize{(a) attributes}}
 \end{center}
\end{minipage}
\begin{minipage}[b]{0.121\linewidth}
 \begin{center}
 \centerline{\includegraphics[width=.9\columnwidth]{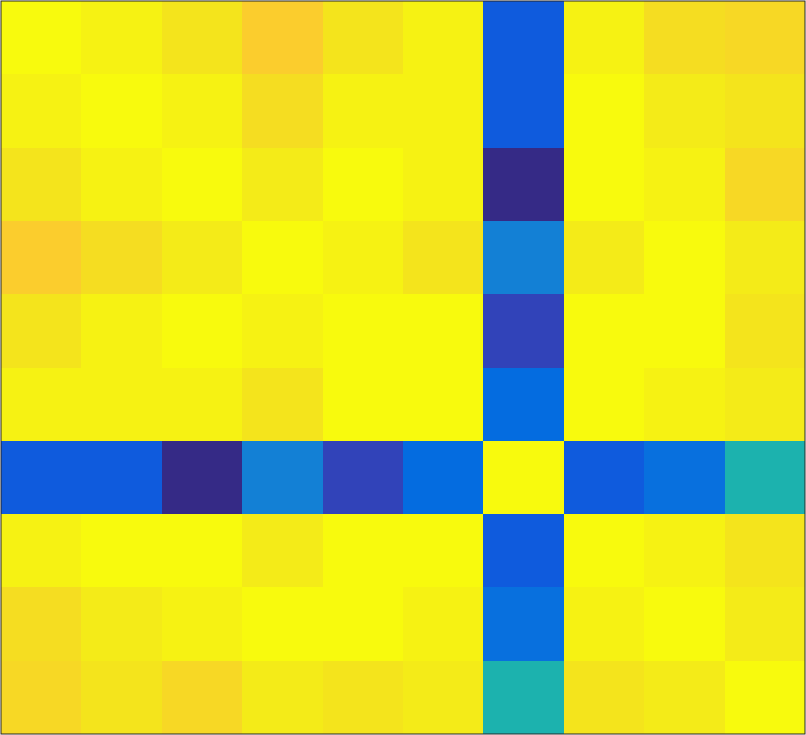}}
 \centerline{\footnotesize{(b) cq-hist (31.5)}}
 \end{center}
\end{minipage}
\begin{minipage}[b]{0.121\linewidth}
 \begin{center}
 \centerline{\includegraphics[width=.9\columnwidth]{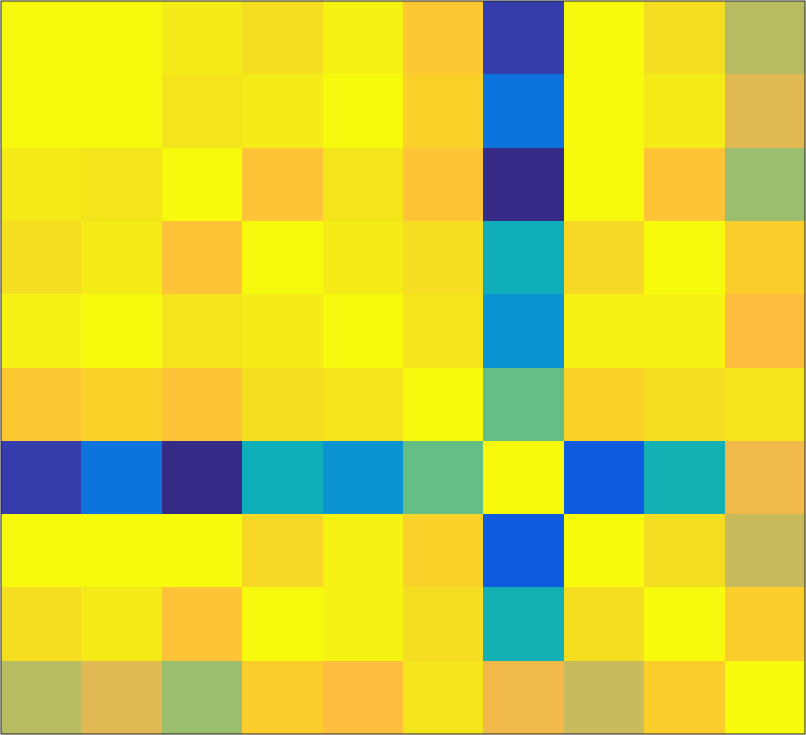}}
 \centerline{\footnotesize{(c) lss-hist (30.3)}}
 \end{center}
\end{minipage}
\begin{minipage}[b]{0.121\linewidth}
 \begin{center}
 \centerline{\includegraphics[width=.9\columnwidth]{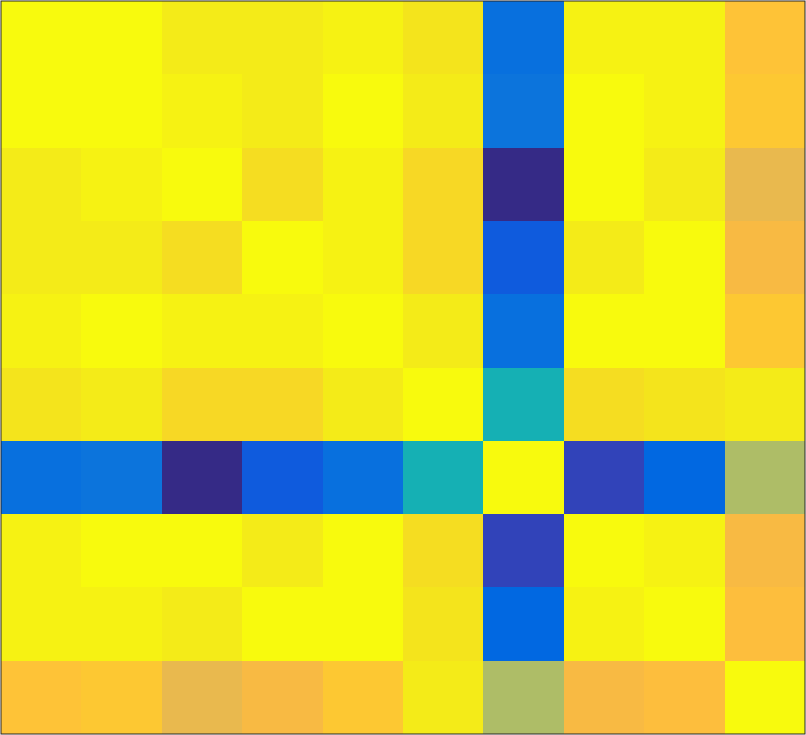}}
 \centerline{\footnotesize{(d) rgsift-hist (33.6)}}
 \end{center}
\end{minipage}
\begin{minipage}[b]{0.121\linewidth}
 \begin{center}
 \centerline{\includegraphics[width=.9\columnwidth]{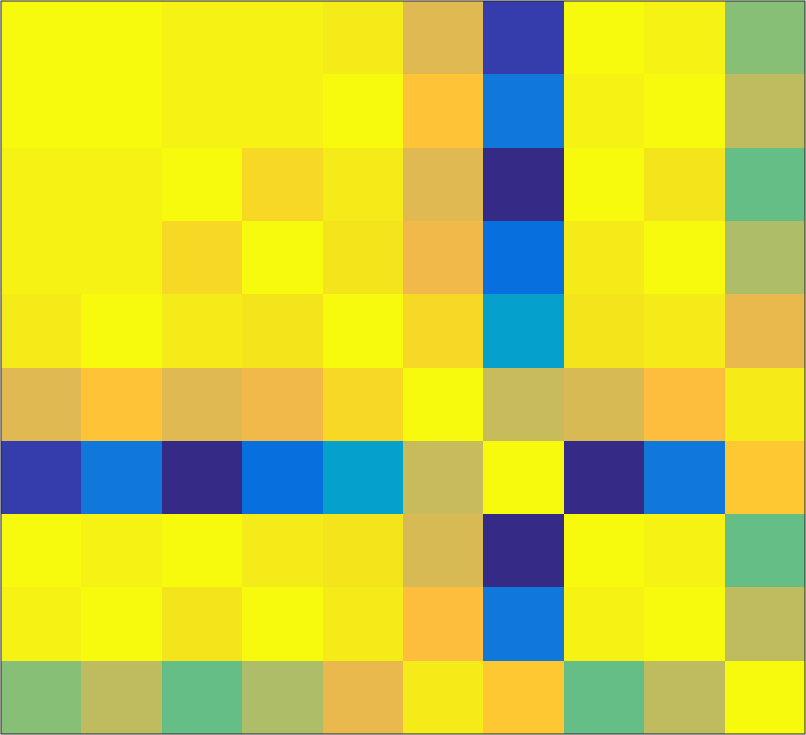}}
 \centerline{\footnotesize{(e) sift-hist (29.8)}}
 \end{center}
\end{minipage}
\begin{minipage}[b]{0.121\linewidth}
 \begin{center}
 \centerline{\includegraphics[width=.9\columnwidth]{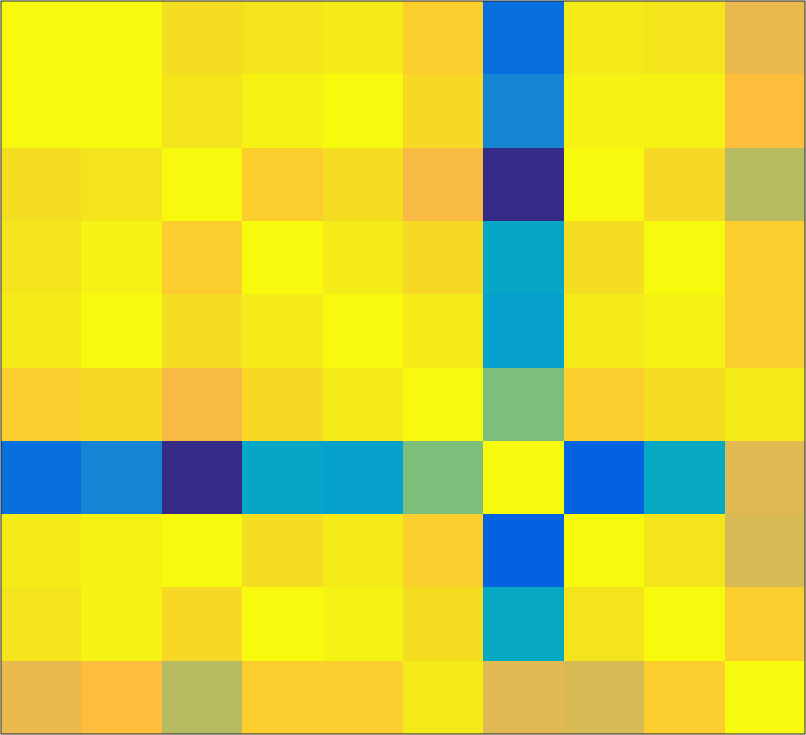}}
 \centerline{\footnotesize{(f) surf-hist (36.5)}}
 \end{center}
\end{minipage}
\begin{minipage}[b]{0.121\linewidth}
 \begin{center}
 \centerline{\includegraphics[width=.9\columnwidth]{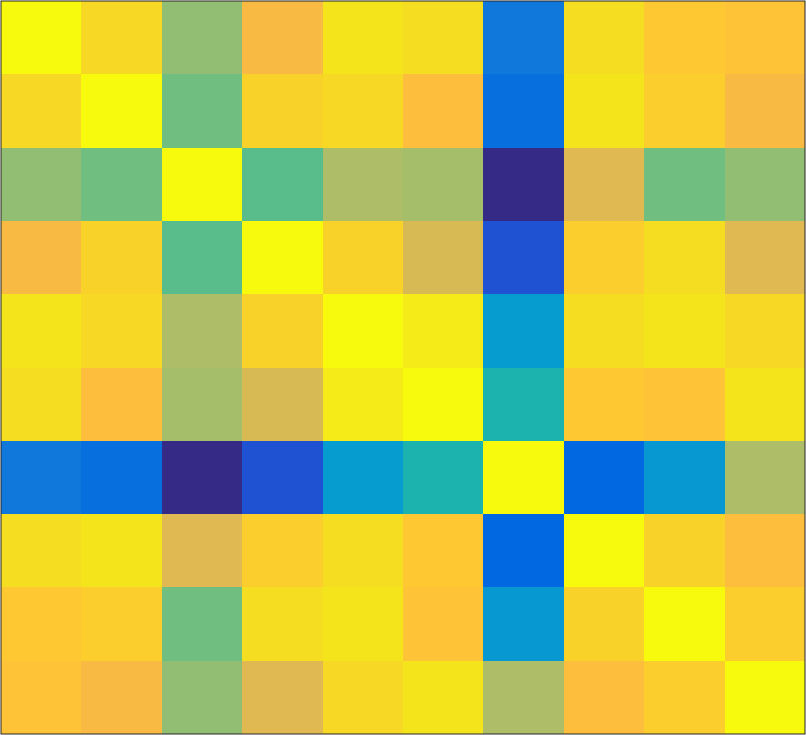}}
 \centerline{\footnotesize{(g) decaf (52.0)}}
 \end{center}
\end{minipage}
\begin{minipage}[b]{0.121\linewidth}
 \begin{center}
 \centerline{\includegraphics[width=.9\columnwidth]{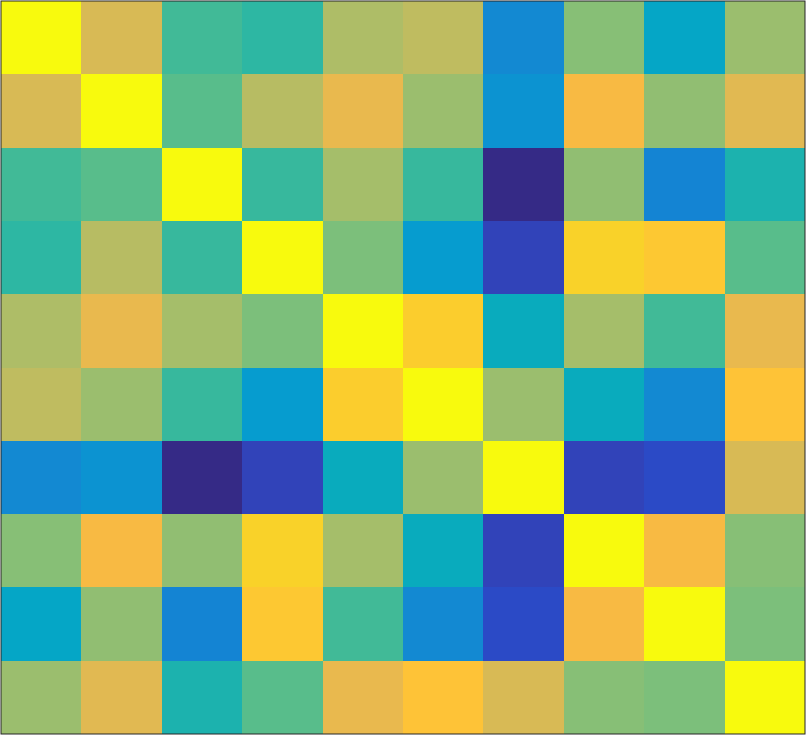}}
 \centerline{\footnotesize{(h) verydeep-19 (71.5)}}
 \end{center}
\end{minipage}
\vspace{-7mm}
\caption{\footnotesize{Cosine similarity matrices created using different features on AwA testing data. The numbers in the brackets are the mean accuray (\%) achieved using the corresponding features. Our learning method performs the best with vgg-verydeep-19 features. We can attribute this to the fact that we need a class dependent feature transform $\phi_y(\mathbf{x})$ that has good separation on seen classes.}}\label{fig:awa-mat}
\vspace{-3mm}
\end{figure*}

For the detail of each dataset, please refer to its original paper. For aP\&Y, CUB-200-2011, and SUN Attribute datasets, we take the means of attribute vectors from the same classes to generate source domain data. For AwA dataset, we utilize the real-number attribute vectors since they are more discriminative.
%
%
%
%

We utilize the same training/testing splits for zero-shot recognition on aP\&Y and AwA as others. For CUB-200-2011, we follow \cite{akata2013label} to use the same 150 bird spices as seen classes for training and the left 50 spices as unseen classes for testing. For SUN Attribute, we follow \cite{jayaraman2014unreliable} to use the same 10 classes as unseen classes for testing (see their supplementary file) and take the rest as seen classes for training.

We summarize our comparison in Table \ref{tab:apy}, where the blank spaces indicate that the proposed methods were not tested on the datasets in their original papers. Still there is no big performance difference between our SSE-INT and SSE-ReLU. On 4 out of the 5 datasets, our method works best except for CUB-200-2011. On one hand, \cite{Akata2015} specifically targets at fine-grained zero-shot recognition such as this dataset, while ours aims for general zero-shot learning. On the other hand, we suspect that the source domain projection function may not work well in fine-grained recognition, and we will investigate more on it in our future work.  

To understand our method better with different features, we test 7 features on AwA dataset\footnote{We downloaded these features from \url{http://attributes.kyb.tuebingen.mpg.de/}}. We show the SSE distribution comparison using decaf CNN features and vgg-verydeep-19 CNN features in Fig. \ref{fig:awa-feat}. There is a large difference between the two distributions: (a) while with decaf features clusters are slightly separated they are still cluttered with overlaps among different classes. (b) vgg-verydeep-19 features, in contrast, form crisp clusters for different classes, which is useful for zero-shot recognition. Also we plot the cosine similarity matrices created using different features in Fig. \ref{fig:awa-mat}. As we see, the matrix from vgg-verydeep-19 features (\ie the last) is the most similar to that from the source domain attribute vectors (\ie the first). This demonstrates that our learning method with vgg-verydeep-19 features can align the target domain distribution with the source domain attribute vectors. We can attribute this to the fact that we need a class dependent feature transform $\phi_y(\mathbf{x})$ that has good separation on seen classes.

Our implementation\footnote{Our code is available at \url{https://zimingzhang.wordpress.com/source-code/}.} is based on unoptimized MATLAB code. However, it can return the prediction results on any of these 5 datasets within 30 minutes using a multi-thread CPU (Xeon E5-2696 v2), starting from loading CNN features. For instance, on CIFAR-10 we manage to finish running the code less than 5 minutes.

\subsection{Towards Large-Scale Zero-Shot Recognition}

\begin{figure}[t]
\begin{minipage}[b]{0.49\columnwidth}
 \begin{center}
 \centerline{\includegraphics[width=\columnwidth]{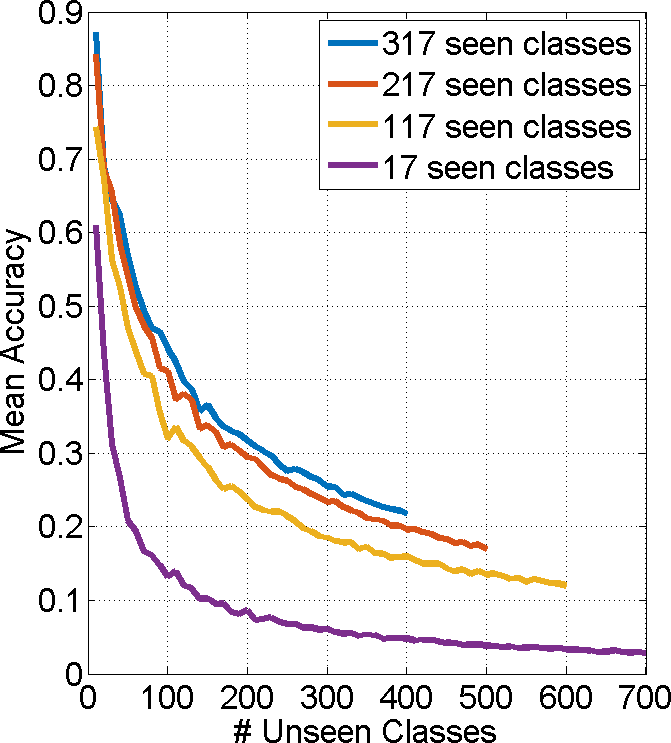}}
 \centerline{\footnotesize{(a) Recognition on unseen classes}}
 \end{center}
\end{minipage}
\begin{minipage}[b]{0.49\columnwidth}
\begin{center}
\centerline{\includegraphics[width=\columnwidth]{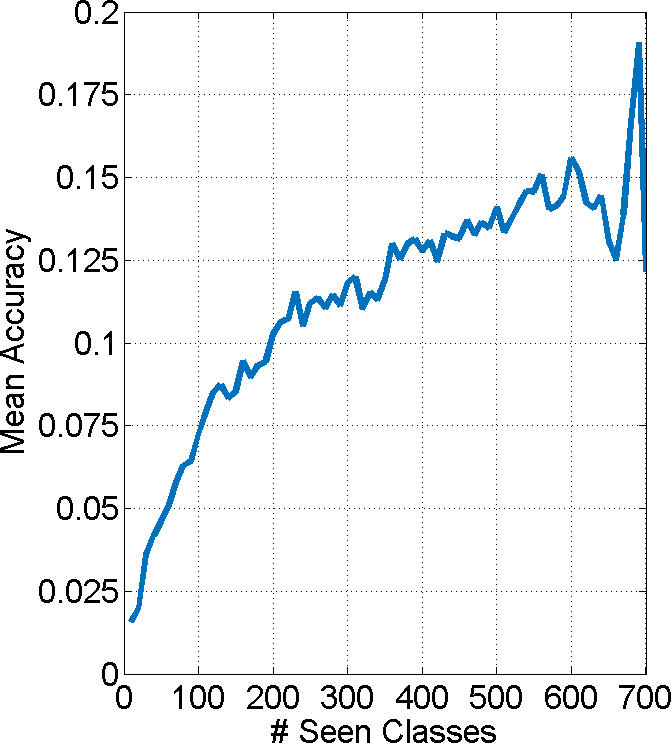}}
 \centerline{\footnotesize{(b) Recognition on all classes}}
\end{center} 
\end{minipage}
\vspace{-3mm}
\caption{\footnotesize{Large-scale zero-shot recognition on SUN Attribute.}}\label{fig:ls}
\vspace{-4mm}
\end{figure}

We test the generalization ability of our method on the SUN Attribute dataset for large-scale zero-shot recognition.
We design two experimental settings: (1) Like in benchmark comparison, we randomly select $M$ classes as seen classes for training, and then among the rest $717-M$ classes, we also randomly select $10, 20, \cdots, 717-M$ classes as unseen classes for testing; (2) We randomly select $10, 20, \cdots, 700$ classes as seen classes for training, and categorize each data sample from the rest unseen classes into one of the 717 classes. Fig. \ref{fig:ls} shows our results, where (a) and (b) correspond to the settings (1) and (2), respectively. 

In Fig. \ref{fig:ls}(a), we can see that with very few seen classes, we can achieve reasonably good performance when unseen classes are a few. However, with the increase of the number of unseen classes, the curve drops rapidly and then changes slowly when the number is large. From 200 to 700 unseen classes, our performance is reduced from 8.62\% to 2.85\%. With the increase of the number of seen classes, our performance is improving, especially when the number of unseen classes is small. With 10 unseen classes, our performance increases from 61.00\% to 87.17\% using 17 and 317 seen classes, respectively. But such improvement is marginal when there are already a sufficient number of seen classes, for instance from 217 to 317 seen classes. 

In Fig. \ref{fig:ls}(b), generally speaking, with more seen classes our performance will be better, because there will be better chance to preserve the semantic affinities among classes in source domain. With only 10 seen classes, our method can achieve 1.59\% mean accuracy, which is much better than the random chance 0.14\%. Notice that even though we use all the 717 classes as seen classes, we cannot guarantee that the testing results are similar to those of traditional classification methods, because the source domain attribute vectors will guide our method for learning. If they are less discriminative, \eg the attribute vectors for cat and dog in CIFAR-10, the recognition performance may be worse. 

To summarize, our method performs well and stably on SUN Attribute with a small set of seen classes and a relatively large set of unseen classes. Therefore, we believe that our method is suitable for large-scale zero-shot recognition.

\section{Conclusion}\label{sec:con}
We proposed learning a semantic similarity embedding (SSE) method for zero-shot recognition. 
We label the semantic meanings using seen classes, and project all the source domain attribute vectors onto the simplex in SSE space, so that each class can be represented as a probabilistic mixture of seen classes. Then we learn similarity functions to embed target domain data into the same semantic space as source domain, so that not only the empirical mean embeddings of the seen class data distributions are aligned with their corresponding source domain embeddings, but also the data instance itself can be classified correctly. We propose learning two variants using intersection function and rectified linear unit (ReLU). Our method on five benchmark datasets including the large-scale SUN Attribute dataset significantly outperforms other state-of-art methods. As future work, we would like to explore other applications for our method such as person re-identification \cite{zhang_eccv14,zhang_iccv15_reid,zhang2014person} and zero-shot activity retrieval \cite{Castanon_mm15}.
%

\section*{Acknowledgement}
We thank the anonymous reviewers for their very useful comments. This material is based upon work supported in part by the U.S. Department of Homeland Security, Science and Technology Directorate, Office of University Programs, under Grant Award 2013-ST-061-ED0001, by ONR Grant 50202168 and US AF contract FA8650-14-C-1728. The views and conclusions contained in this document are those of the authors and should not be interpreted as necessarily representing the social policies, either expressed or implied, of the U.S. DHS, ONR or AF.


{\footnotesize
\bibliographystyle{ieee}
\bibliography{egbib}

\begin{thebibliography}{10}\itemsep=-1pt

\bibitem{akata2013label}
Z.~Akata, F.~Perronnin, Z.~Harchaoui, and C.~Schmid.
\newblock Label-embedding for attribute-based classification.
\newblock In {\em CVPR}, pages 819--826, 2013.

\bibitem{Akata2015}
Z.~Akata, S.~Reed, D.~Walter, H.~Lee, and B.~Schiele.
\newblock Evaluation of output embeddings for fine-grained image
  classification.
\newblock In {\em CVPR}, June 2015.

\bibitem{bengio2010label}
S.~Bengio, J.~Weston, and D.~Grangier.
\newblock Label embedding trees for large multi-class tasks.
\newblock In {\em NIPS}, pages 163--171, 2010.

\bibitem{Berg:2010:AAD:1886063.1886114}
T.~L. Berg, A.~C. Berg, and J.~Shih.
\newblock Automatic attribute discovery and characterization from noisy web
  data.
\newblock In {\em ECCV}, pages 663--676, 2010.

\bibitem{blanchard2014decontamination}
G.~Blanchard and C.~Scott.
\newblock Decontamination of mutually contaminated models.
\newblock In {\em AISTATS}, 2014.

\bibitem{Castanon_mm15}
G.~D. Castanon, Y.~Chen, Z.~Zhang, and V.~Saligrama.
\newblock Efficient activity retrieval through semantic graph queries.
\newblock In {\em ACM Multimedia}, 2015.

\bibitem{elhamifar2013sparse}
E.~Elhamifar and R.~Vidal.
\newblock Sparse subspace clustering: Algorithm, theory, and applications.
\newblock {\em PAMI}, 35(11):2765--2781, 2013.

\bibitem{farhadi2009attribute}
A.~Farhadi, I.~Endres, D.~Hoiem, and D.~Forsyth.
\newblock Describing objects by their attributes.
\newblock In {\em CVPR}, pages 1778--1785, 2009.

\bibitem{frome2013devise}
A.~Frome, G.~S. Corrado, J.~Shlens, S.~Bengio, J.~Dean, M.~A. Ranzato, and
  T.~Mikolov.
\newblock Devise: A deep visual-semantic embedding model.
\newblock In {\em NIPS}, pages 2121--2129, 2013.

\bibitem{embedding2014ECCV}
Y.~Fu, T.~M. Hospedales, T.~Xiang, Z.~Fu, and S.~Gong.
\newblock Transductive multi-view embedding for zero-shot recognition and
  annotation.
\newblock In {\em ECCV}, 2014.

\bibitem{DBLP:journals/corr/HammB15}
J.~Hamm and M.~Belkin.
\newblock Probabilistic zero-shot classification with semantic rankings.
\newblock {\em CoRR}, abs/1502.08039, 2015.

\bibitem{hwang2013analogy}
S.~J. Hwang, K.~Grauman, and F.~Sha.
\newblock Analogy-preserving semantic embedding for visual object
  categorization.
\newblock In {\em ICML}, pages 639--647, 2013.

\bibitem{hwang2014unified}
S.~J. Hwang and L.~Sigal.
\newblock A unified semantic embedding: Relating taxonomies and attributes.
\newblock In {\em NIPS}, pages 271--279, 2014.

\bibitem{jayaraman2014unreliable}
D.~Jayaraman and K.~Grauman.
\newblock Zero-shot recognition with unreliable attributes.
\newblock In {\em NIPS}, pages 3464--3472, 2014.

\bibitem{citeulike:7491128}
A.~Krizhevsky.
\newblock {Learning Multiple Layers of Features from Tiny Images}.
\newblock Master's thesis, 2009.

\bibitem{10.1109/TPAMI.2013.140}
C.~H. Lampert, H.~Nickisch, and S.~Harmeling.
\newblock Attribute-based classification for zero-shot visual object
  categorization.
\newblock {\em PAMI}, 36(3):453--465, 2014.

\bibitem{liu2013robust}
G.~Liu, Z.~Lin, S.~Yan, J.~Sun, Y.~Yu, and Y.~Ma.
\newblock Robust recovery of subspace structures by low-rank representation.
\newblock {\em PAMI}, 35(1):171--184, 2013.

\bibitem{mahajan2011joint}
D.~Mahajan, S.~Sellamanickam, and V.~Nair.
\newblock A joint learning framework for attribute models and object
  descriptions.
\newblock In {\em ICCV}, pages 1227--1234, 2011.

\bibitem{maji2008classification}
S.~Maji, A.~C. Berg, and J.~Malik.
\newblock Classification using intersection kernel support vector machines is
  efficient.
\newblock In {\em CVPR}, pages 1--8, 2008.

\bibitem{mensink2014costa}
T.~Mensink, E.~Gavves, and C.~G.~M. Snoek.
\newblock Costa: Co-occurrence statistics for zero-shot classification.
\newblock In {\em CVPR}, pages 2441--2448, June 2014.

\bibitem{mensink2012metric}
T.~Mensink, J.~Verbeek, F.~Perronnin, and G.~Csurka.
\newblock Metric learning for large scale image classification: Generalizing to
  new classes at near-zero cost.
\newblock In {\em ECCV}, pages 488--501. 2012.

\bibitem{nair2010rectified}
V.~Nair and G.~E. Hinton.
\newblock Rectified linear units improve restricted boltzmann machines.
\newblock In {\em ICML}, pages 807--814, 2010.

\bibitem{norouziMBSSFCD14}
M.~Norouzi, T.~Mikolov, S.~Bengio, Y.~Singer, J.~Shlens, A.~Frome, G.~S.
  Corrado, and J.~Dean.
\newblock {Zero-Shot Learning by Convex Combination of Semantic Embeddings}.
\newblock In {\em ICLR}, 2014.

\bibitem{palatucci2009zero}
M.~Palatucci, D.~Pomerleau, G.~E. Hinton, and T.~M. Mitchell.
\newblock Zero-shot learning with semantic output codes.
\newblock In {\em NIPS}, pages 1410--1418, 2009.

\bibitem{Parikh:2011:IBD:2191740.2191861}
D.~Parikh and K.~Grauman.
\newblock Interactively building a discriminative vocabulary of nameable
  attributes.
\newblock In {\em CVPR}, pages 1681--1688, 2011.

\bibitem{patterson2014sun}
G.~Patterson, C.~Xu, H.~Su, and J.~Hays.
\newblock The sun attribute database: Beyond categories for deeper scene
  understanding.
\newblock {\em IJCV}, 108(1-2):59--81, 2014.

\bibitem{peng2012constructing}
X.~Peng, L.~Zhang, and Z.~Yi.
\newblock Constructing l2-graph for subspace learning and segmentation.
\newblock {\em arXiv preprint arXiv:1209.0841}, 2012.

\bibitem{conf/nips/RohrbachES13}
M.~Rohrbach, S.~Ebert, and B.~Schiele.
\newblock Transfer learning in a transductive setting.
\newblock In {\em NIPS}, pages 46--54, 2013.

\bibitem{rohrbach2011largeScale}
M.~Rohrbach, M.~Stark, and B.~Schiele.
\newblock Evaluating knowledge transfer and zero-shot learning in a large-scale
  setting.
\newblock In {\em CVPR}, pages 1641--1648, 2011.

\bibitem{Romera-Paredes2015}
B.~Romera-Paredes and P.~H.~S. Torr.
\newblock An embarrassingly simple approach to zero-shot learning.
\newblock In {\em ICML}, 2015.

\bibitem{ILSVRCarxiv14}
O.~Russakovsky, J.~Deng, H.~Su, J.~Krause, S.~Satheesh, S.~Ma, Z.~Huang,
  A.~Karpathy, A.~Khosla, M.~Bernstein, A.~C. Berg, and L.~Fei-Fei.
\newblock {ImageNet Large Scale Visual Recognition Challenge}, 2014.

\bibitem{simonyan2014very}
K.~Simonyan and A.~Zisserman.
\newblock Very deep convolutional networks for large-scale image recognition.
\newblock {\em arXiv preprint arXiv:1409.1556}, 2014.

\bibitem{smola2007hilbert}
A.~Smola, A.~Gretton, L.~Song, and B.~Sch{\"o}lkopf.
\newblock A hilbert space embedding for distributions.
\newblock In {\em Algorithmic Learning Theory}, pages 13--31, 2007.

\bibitem{socher2013zero}
R.~Socher, M.~Ganjoo, C.~D. Manning, and A.~Ng.
\newblock Zero-shot learning through cross-modal transfer.
\newblock In {\em NIPS}, pages 935--943, 2013.

\bibitem{van2008visualizing}
L.~Van~der Maaten and G.~Hinton.
\newblock Visualizing data using {t-SNE}.
\newblock {\em JMLR}, 9(2579-2605):85, 2008.

\bibitem{arXiv:1412.4564}
A.~Vedaldi and K.~Lenc.
\newblock Matconvnet -- convolutional neural networks for {MATLAB}.
\newblock {\em CoRR}, abs/1412.4564, 2014.

\bibitem{vidal2010tutorial}
R.~Vidal.
\newblock A tutorial on subspace clustering.
\newblock {\em Signal Processing Magazine}, pages 52--68, 2010.

\bibitem{WahCUB_200_2011}
C.~Wah, S.~Branson, P.~Welinder, P.~Perona, and S.~Belongie.
\newblock {The Caltech-UCSD Birds-200-2011 Dataset}.
\newblock Technical report, 2011.

\bibitem{wang2013unified}
X.~Wang and Q.~Ji.
\newblock A unified probabilistic approach modeling relationships between
  attributes and objects.
\newblock In {\em ICCV}, pages 2120--2127, 2013.

\bibitem{Weinberger08large}
K.~Weinberger and O.~Chapelle.
\newblock Large margin taxonomy embedding for document categorization.
\newblock In {\em NIPS}, pages 1737--1744. 2009.

\bibitem{wu2014zero}
S.~Wu, S.~Bondugula, F.~Luisier, X.~Zhuang, and P.~Natarajan.
\newblock Zero-shot event detection using multi-modal fusion of weakly
  supervised concepts.
\newblock In {\em CVPR}, pages 2665--2672, 2014.

\bibitem{yu2013designing}
F.~X. Yu, L.~Cao, R.~S. Feris, J.~R. Smith, and S.~F. Chang.
\newblock Designing category-level attributes for discriminative visual
  recognition.
\newblock In {\em CVPR}, pages 771--778, 2013.

\bibitem{yuille2003concave}
A.~L. Yuille and A.~Rangarajan.
\newblock The concave-convex procedure.
\newblock {\em Neural computation}, 15(4):915--936, 2003.

\bibitem{zhang_eccv14}
Z.~Zhang, Y.~Chen, and V.~Saligrama.
\newblock A novel visual word co-occurrence model for person re-identification.
\newblock In {\em ECCV Workshop on Visual Surveillance and Re-Identification},
  2014.

\bibitem{zhang_iccv15_reid}
Z.~Zhang, Y.~Chen, and V.~Saligrama.
\newblock Group membership prediction.
\newblock In {\em ICCV}, 2015.

\bibitem{zhang2014person}
Z.~Zhang and V.~Saligrama.
\newblock {PRISM}: Person re-identification via structured matching.
\newblock {\em arXiv preprint arXiv:1406.4444}, 2014.

\end{thebibliography}
}

\end{document}